# A novel knowledge graph development for industry design: A case study on indirect coal liquefaction process

***Abstract*:** Hazard and operability analysis (HAZOP) is a remarkable representative in industrial safety engineering. However, a great storehouse of industrial safety knowledge (ISK) in HAZOP reports has not been thoroughly exploited. In order to reuse and unlock the value of ISK and optimize HAZOP, we have developed a novel knowledge graph for industrial safety (ISKG) with HAZOP as the carrier through bridging data science and engineering design. Specifically, firstly, considering that the knowledge contained in HAZOP reports of different processes in industry is not the same, we creatively develope a general ISK standardization framework, it provides a practical scheme for integrating HAZOP reports from various processes and uniformly representing the ISK with diverse expressions. Secondly, we conceive a novel and reliable information extraction model based on deep learning combined with data science, it can effectively mine ISK from HAZOP reports, which alleviates the obstacle of ISK extraction caused by the particularity of HAZOP text. Finally, we build ISK triples and store them in the Neo4j graph database. We take indirect coal liquefaction process as a case study to develop ISKG, and its oriented applications can optimize HAZOP and mine the potential of ISK, which is of great significance to improve the security of the system and enhance prevention awareness for people. ISKG containing the ISK standardization framework and the information extraction model sets an example of the interaction between data science and engineering design, which can enlighten other researchers and extend the perspectives of industrial safety.

## *1. INTRODUCTION*

Hazard and operability analysis (HAZOP) is a prevailing risk analysis method in industry, which can provide safety analysis decision-making for any process in the industry. HAZOP can predict the spread of hazard events through the potential deviation of nodes in the system, and put forward effective solutions. The analysis results are finally recorded in the HAZOP report in the form of text (Kang and Guo, 2016; Dunjo et al., 2010), which makes the HAZOP report contain a great storehouse of industrial safety knowledge (ISK). Yet there are the following weaknesses in existing HAZOP researches:

(1) HAZOP is an expert-driven engineering design in the form of brainstorming (Baybutt, 2015), which relies on the experience and execution of analysts and lasts for at least several weeks, which makes HAZOP time-consuming and labor-consuming, and the analysis results may not be comprehensive.

(2) For the same process, various HAZOP reports will emerge in analysis strategies of different expert teams (Cameron et al., 2017), let alone different processes, which hinders the integration, sharing and reuse of ISK. The release of the ISK value is in a dilemma.

For the weakness (1), plenty of engineering designs have emerged to improve the HAZOP efficiency (Single, Schmidt and Denecke, 2019; Single, Schmidt and Denecke, 2020; Rodriguez et al., 2012; Venkatasubramanian, Zhao and Viswanathan, 2000). For the oil and gas industry, Wu and Lind (2018) proposed multilevel flow modeling framework (MFM) to simulate and reason processes, which can improve HAZOP quality and efficiency with low labor cost. Pasman et al. (2016) introduced system engineering and artificial intelligence into HAZOP, discussed the possibility and cost of improving the HAZOP efficiency, and identified the causality in risk accidents by constructing ontology. Taking glycerol nitration process and ammonia synthesis process as case studies, Janošovský et al. (2017) combined HAZOP with Aspen HYSYS (a process simulator) to improve the performance of hazard recognizer. Ramzan et al. (2007) extended HAZOP by introducing dynamic simulation and risk matrix, built a steady-state simulation model and a dynamic simulation model based on the distillation column unit. Jeerawongsuntorn, Sainyamsatit and Srinophakun (2011) considered HMI (a human-machine interface) to automatically manage HAZOP based on a safety instrument system, which successfully reduced the time of hazard identification in the biodiesel production process. In addition, the computer aided HAZOP based on mathematical calculation and P & ID information (Lü and Wang, 2007), the industrial assistant (IRIS) for product life cycle management system (Bragatto, 2007), and the safety analysis system with natural language processing technology (Daramola et al., 2013), etc. These computer-based or automation-assisted studies have optimized HAZOP (Feng et al., 2008).

For the weakness (2), with the rise of artificial intelligence, some researches have devoted to introducing natural language processing technology into HAZOP reports to mine semantic information and ISK in recent years. For instance, Feng et al. (2021) classified the consequence severity level of HAZOP reports based on deep learning, which is helpful for small-scale chemical plants in safety analysis. Wang, Zhang, and Gao (2021) designed three novel active learning algorithms to build the entity recognition model and mine the ISK in HAZOP reports, which is of great significance to improve industrial safety. Hu et al. (2015) proposed a fault diagnosis method of integrating HAZOP and dynamic bayesian network reasoning, which can reveal the early deviation in the fault causal chain and play the role in emergency decision-making.

Unfortunately, the above-mentioned researches are mainly to alleviate weaknesses (1), and lacks a profound study in weaknesses (2). Therefore, what is in urgent demand is a feasible and advanced approach that can surmount the two problems concurrently. In this research, inspired by the knowledge exploration in engineering design (Chiarello,

Fantoni and Bonaccorsi, 2017; Chiarello et al., 2019; Fantoni et al., 2021), we have developed a novel knowledge graph (KG) (Sidor, 2015) with HAZOP as the carrier for industrial safety through bridging data science and engineering design, named ISKG, which can unlock the value of ISK, and improve the efficiency and comprehensiveness of HAZOP. Specifically, Firstly, considering that the knowledge contained in HAZOP reports is diverse, we have creatively developed a general ISK standardization framework (ISKSF) by deconstructing and generalizing HAZOP in a top-down manner. ISKSF contains the ISK ontology and the ISK relationship, which can standardize HAZOP reports in various processes, uniformly represent and integrate different types of ISK, it is a practice that broadens the research of engineering design for industrial safety. Secondly, considering the particularity of HAZOP text, we skillfully conceive a novel and reliable information extraction model (HAINEX) based on deep learning in combination with data science, HAINEX can extract the ISK in HAZOP reports based on the ISKSF, which is a practical application that can extend the perspective of data science in engineering design about the industrial information with strong structure and logic. Briefly, HAINEX consists of three modules: an optimized pre-training language model termed IBERT for extracting semantic features, an encoder for obtaining the context features through the bidirectional long short-term memory network (BiLSTM) (Hochreiter et al., 1997; Lindemann et al., 2021), and a decoder based on conditional random field (CRF) (Sutton and Mccallum, 2006; Zhang et al., 2018) with an improved industrial loss function termed IL. HAINEX improves the efficiency of ISK extraction by treating features as a candidate set and screening them. Finally, we construct ISK triples and store them in the Neo4j graph database based on ISKSF and HAINEX. We cooperated with Shenhua Group (Nolan, Shipman and Rui, 2004) to launch HAZOP and develop ISKG for the indirect coal liquefaction process. ISKG can integrate, share and reuse ISK, and unlock the value of HAZOP reports, it is an example of the interaction between data science and engineering design that can inspire other researchers. ISKG oriented applications, such as ISK visualization, ISK retrieval, auxiliary HAZOP and hazard propagation reasoning, can mine the potential of ISK, optimize HAZOP and improve HAZOP execution efficiency, which is of great significance to improve the security of the system. What's more, the ISKG based question-answering system can provide teaching guidance to popularize the safety knowledge and enhance prevention awareness for non-professionals. These applications that inject data science into engineering design can be used as the support for the industrial development. The main contributions of this research are as follows.

(1) We have developed ISKG with HAZOP as the carrier through bridging data science and engineering design, which can reuse and unlock the value of ISK, and improve the efficiency and comprehensiveness of HAZOP.

(2) The proposed ISKSF can provide a practical scheme for standardizing different HAZOP reports and uniformly representing the ISK with diverse expressions, which can realize the integration and circulation of ISK.

(3) The novel and reliable HAINEX can effectively extract ISK from HAZOP reports, which provides an optimized idea and feasible scheme for knowledge extraction in industrial information with strong structure and logic.

(4) We have launched ISKG for indirect coal liquefaction process, its oriented applications are of great significance in optimizing HAZOP, strengthening industrial safety and popularizing safety prevention awareness.

Section 2 is the related work, which mainly reviews HAZOP and KG. Section 3 is the method, we elaborate the development process of ISKG, mainly including ISKSF and HAINEX. In Section 4, we present the indirect coal liquefaction process, demonstrate ISKG and its various applications. The conclusion is expounded in Section 5. In addition, since the HAZOP report is in Chinese, we have given the corresponding translation in English to the examples involved, the sentences structured as " description in Chinese (description in English)".

## *2. RELATED WORK*

In this paper, we can skillfully unify HAZOP and KG, hence, we launch related work on HAZOP and KG, respectively.

### *2.1. HAZOP*

HAZOP is an analysis process with inverse reasoning (Panagiotis et al., 2019), its main purpose is to identify and explore potential facility hazards and mis-operations. Experts take the potential deviation (such as the change of unit nature) in the industry as the breakthrough point, with the assistance of guide words (such as "伴随(accompany)", "相反(opposite)" and "异常(abnormal)"), to deduce accident scenarios in industrial operation logically, assess risk level in predicted accidents, and formulate measures of curbing hazards, which prevents the occurrence of accidents (Li, 2020). It is worth noting that the industrial system needs to be divided into multiple nodes to facilitate HAZOP (EPSC, 2000). Generally, for the continuous operation, the node is the process unit, such as "费托反应器(Fischer Tropsch reactor)", and for the intermittent operation, the node is the operation procedure, such as "光气和多胺反应生产多亚甲基多苯基多异氰酸酯(polymethylene polyphenyl polyisocyanate was produced by the reaction of phosgene and polyamine)". The size of nodes is related to system complexity and hazard severity.

At present, HAZOP has been widely implemented in various processes and achieved superb performance. For example, in the food industry, Lee et al. (2021) took the outbreak of listeria monocytogenes in the United States as a case study and analyzed and improved the standards for planning, design and implementation in the food safety management system with the help of HAZOP. In the palm oil production industry, Lim et al. (2021) leveraged HAZOP to conduct a detailed safety review of the current palm oil process, evaluate and adaptively correct the potential implementation deficiencies of industry 4.0 technology. In the organic hydride hydrogen refueling process, Suzuki, Izato, and Miyake (2021) simulated the accident scene caused by the ignition of internal factors of the organic hydride hydrogenation station in the HAZOP environment, which improved the staff's prevention of hazards. In the formic acid process, Patle et al. (2021) proposed a complete set of process design, control and safety analysis methods based on the dynamic simulation of HAZOP, which improved the safety of the process. Furthermore, HAZOP also plays an irreplaceable role in risk assessment and safety optimization, such as, the Thai Research Reactor-1/Modification 1 (Vechgama et al., 2021), Oil facilities in the Chamshir Dam water basin (Galalizadeh et al., 2020), the dismantling of large industrial machines and associated structures (Joubert, Steyn and Pretorius, 2020) and the multizone circulating reactor (He et al., 2020). Obviously, HAZOP has penetrated into the industrial field, for the most critical analysis results that contain valuable yet heterogeneous ISK, the HAZOP reports, what is urgently carried out is a research that can effectively manage them.

### 2.2. KG

The KG is a semantic network composed of < head, relation, tail > triples in the form of directed graph (Sowa, 2014; Sidor, 2015), which describes ontologies and the relationship between them in a standardized way. In the era of information explosion, KG, a promising engine, has bright prospects in eliciting, integrating, processing, utilizing and popularizing mass data embedded in services (Hao et al., 2021; Zhou et at., 2021). Construction period, deduction period and using period are roughly three stages of developing KG, where the construction period is basic and essential (Li et at., 2021). Ontology design (Fatfouta and Le-Cardinal, 2021) and data filling (Gesese et al., 2020) are the two linchpins in the construction period, since ontology is the "skeleton" that indicates the type and prospect of KG, data is the "blood" to determine the hierarchy, richness and extensibility of KG. The deduction period and the using period are commonly intertwined and widely exist in downstream tasks such as question-answering system (Phan and Do, 2021), retrieval system (Zhang et al., 2013), information visualization (Cox et al., 2020) and knowledge recommendation (Sacenti, Fileto and Willrich, 2021).

The KG has gained widespread concern due to its capability of storing and sharing knowledge, semi-structured and unstructured, elicited from heterogeneous domains, such as DBpedia KG (Bizer et al., 2009) of general encyclopedia domain and its Chinese version (Xu et al., 2017). Furthermore, KGs have been developed and applied in vertical fields, such as COVID-19 diagnosis (Zheng et al., 2021), RNA analysis of cells (Doddahonnaiah et al., 2021) and film emotion evaluation (Arno et al., 2021), etc. (Xiong et al., 2021; Mawkhiew, Sahoo and Kharshiing. 2021). Since industry tends to be more practical (Chiarello, Belingheri and Fantoni, 2021), in recent years, researchers have attempted to drive industrial services with KG to optimize process safety and product quality (Yahya et al., 2021). Liu et al. (2021) utilized KG to explore potential rules in railway faults and put forward preventive measures. Sarazin et al. (2021) combined KG with aviation system, proposed an advanced expert system for heterogeneous information network, and developed condition maintenance of complex system. Guo et al. (2021) proposed a framework for automatic construction of knowledge base in machining field based on KG, which overcomes the shortcomings of time-consuming and labor-consuming in traditional frameworks. Zhou et al. (2021) constructed a workshop resource KG to integrate the engineering information of the processing workshop, and proposed a unified KG driven production resource allocation method, which allows rapid resource allocation decisions to order tasks. Huet et al. (2021) combined design rules and context information to build a computable KG, which further improved the ability of computer-aided design and allowed designers to spend more time on design rather than looking for design rules. These wonderful cases demonstrate the great potential of KG in industry.

## 3. METHOD

In this section, we elaborate on ISKG development with HAZOP as the carrier, which is also the process of applying data science to engineering design. Firstly, considering that the knowledge contained in HAZOP reports of different processes in industry is not the same, we creatively developed a general ISK standardization framework (ISKSF) to standardize HAZOP reports in various processes, uniformly represent and integrate different types of ISK, which is called "pattern layer". Secondly, considering the particularity of HAZOP text, we skillfully conceive the HAINEX model based on deep learning in combination with data science, which can extract the ISK in HAZOP reports based on the ISKSF, which is called "data layer". Finally, we structure ISK triples and store them in the Neo4j graph database, which is called " storage layer". Fig. 1 is the construction of ISKG. It is worth noting that the pattern layer and the data layer are the focus, since they are directly related to the quality of ISKG, moreover, using natural language processing to explore massive unstructured or semi-structured data is a meaningful and promising challenge (Chiarello, Belingheri and Fantoni, 2021). The following is a detailed introduction.

### 3.1. Pattern layer

The knowledge contained in HAZOP reports of different processes in industry is not the same, such as the electronic agile radar and the WS11 turbofan engine in the aviation industry, the Fischer Tropsch reactor and the circulating gas compressor in the coal industry, etc. Moreover, even for the same process, HAZOP descriptions analyzed by different expert teams are various. Consequently, the design of ISK ontology and its relationship faces great challenges. In view of this difficulty, we propose the ISKSF to standardize the HAZOP information of the same process or even different processes, and uniformly represent different types of ISK. In addition, we design a novel and feasible relationship for the circulation of ISK. Details are as follows.

Due to multitudinous obscure professional terms in HAZOP reports, abstract methods such as induction and clustering are poorly suitable for designing the ISK ontology. Hence, we attempt to consider the nature of HAZOP as a breakthrough. Inspired by the international standard IEC 61882 (IEC, 2001), we use the top-down approach to methodically deconstruct HAZOP. Firstly, HAZOP is divided into a variety of hazard propagation events (HEs), then, the HE is dissected into common elements, and finally the ISK ontology is obtained.

We disassemble HAZOP into HEs with set U = {Cause, Deviation, Middle event, Consequence, Suggestion}. The elements in U are briefly described below.

(1) Cause (IC), the cause of D, such as "管线存水(pipeline water storage)" and "员工误操作(staff mis-operation)".

(2) Deviation (D), the entry point for implementing HAZOP, is mainly composed of guiding words and parameters. The common guiding words are "伴随(accompaniment)", "反向(reverse)" and "异常(abnormal)", etc. The common parameters are "液位(liquid level)", "氧含量(oxygen content)", "温度(temperature)" and "流量(flow)", etc.

(3) Middle event (ME), the successive abnormal reaction of units. For example, "阀门开度过小(too small valve opening)", "液位过低(too low liquid level)", "温度过高(too high temperature)" and "压力过大(too high pressure)" are a series of ME. It is worth noting that a HE may contain more than one ME.

(4) Consequence (C), the negative impact caused by D, such as fire, casualties and release of harmful substances, etc. The type of C is determined by the IC and ME.

(5) Suggestion (S), the risk prevention, such as, "增加液位报警装置(adding liquid level alarm device)".

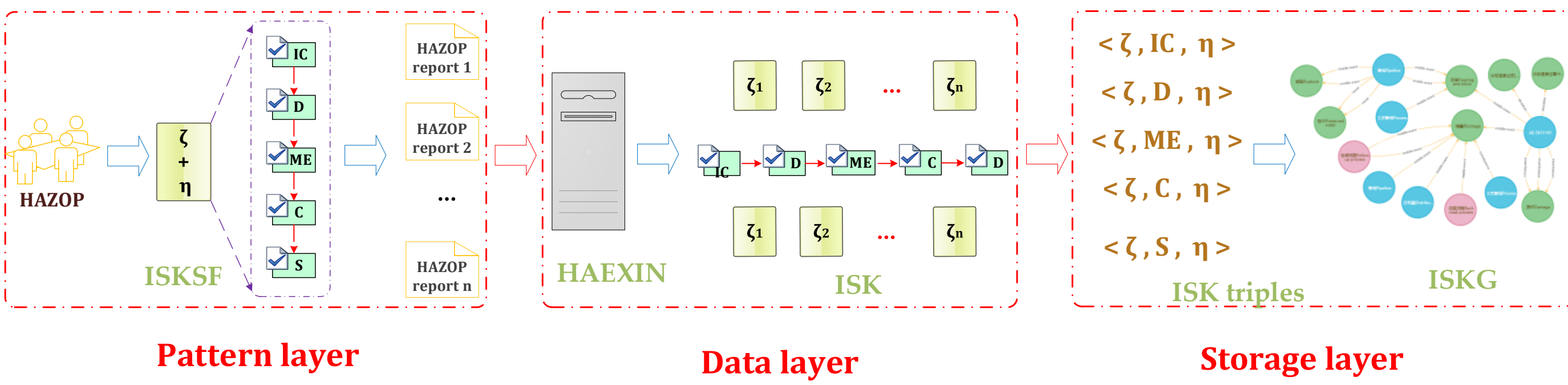

**Fig. 1.** The ISKG development

According to the discrepancy between elements in U, we appropriately divide HEs into single string HE ($E_{ss}$), reverse tree HE ($E_{rt}$), positive tree HE ($E_{pt}$) and bow tie tree HE ($E_{bt}$), as shown in definition (1).

$$HE \in \{E_{ss}, E_{rt}, E_{pt}, E_{bt}\} \tag{1}$$

Where, $E_{ss}$ is the simplest HE with single causal chain. For $E_{rt}$, the same C is caused by various ICs, and how to exhaust all ICs is extremely knotty in analyzing $E_{rt}$. For $E_{pt}$, each ME will engender multiple Cs, which obviously brings more obstacles to subsequent protective measures. $E_{bt}$ is the most intricate, which integrates $E_{ss}$, $E_{rt}$ and $E_{pt}$. Generally, the process units are closely connected, if one unit fails, other units will be involved. Hence, $E_{bt}$ is a common HE in industrial safety, it has attracted extensive attention of engineers. Equations (2) - (5) are their definitions.

$$E_{ss} = \left\{IC, D, \sum ME_i, C, S\right\} \tag{2}$$

$$E_{rt} = \left\{\sum IC_i, D, \sum ME_j, C, S\right\} \tag{3}$$

$$E_{pt} = \left\{IC, D, \sum ME_i, \sum C_j, S\right\} \tag{4}$$

$$E_{bt} = \left\{\sum IC_i, D, \sum ME_j, \sum C_k, S\right\} \tag{5}$$

Furthermore, we generalize each element in each HE into the form of "ζ + η". Where, ζ represents the equipment in the process, the external material or the generated material, and η represents the operating state of the equipment or the physical state of materials. ζ and η are already the basic knowledge. Now, we have obtained a general framework "ζ + η" that can be used to represent HAZOP, and take it as the ISK ontology ($O_{ISK}$), as shown in Equ.6. We take $E_{ss}$ (ζ, η) (see Equ.7) generalized by $E_s$ as an example to expand a brief introduction, as shown in Table 1. It is worth noting that $O_{ISK}$ is built based on HAZOP, hence it is applicable to any process under the HAZOP system, the description in HAZOP reports of different processes can be represented by $O_{ISK}$ and then unified.

$$O_{ISK} = E_{ss}(\zeta,\eta) \cup E_{rt}(\zeta,\eta) \cup E_{pt}(\zeta,\eta) \cup E_{bt}(\zeta,\eta) \tag{6}$$

$$E_{ss}(\zeta,\eta) = \left\{IC(\zeta,\eta), D(\zeta,\eta), \sum ME_i(\zeta,\eta), C(\zeta,\eta), S(\zeta,\eta)\right\} \tag{7}$$

More importantly, different from the general KG relationship, there is no general relationship in sense between ISK in HAZOP descriptions. For example, for a general factual description, "马云创立阿里巴巴(Jack Ma founded Alibaba)", it is obvious that the relationship between "马云(Jack Ma)" and "阿里巴巴(Alibaba)" is "建立(founded)", while for a HAZOP description, "反应堆裂变(reactor fission)", we can observe that no corresponding sense relationship can be found between "反应堆(reactor)" and "裂变(fission)". To surmount this obstacle, we should jump out of the shackles of the traditional KG. There are two considerations.

(1) ISK relationship aims to connect ISK to form an organic whole, the HAZOP description consists of ζ and η. ISK relationship can be equivalent to the relationship between ζ and η with HAZOP as the carrier.

(2) HAZOP is executed based on the criterion of "the flow of the node deviation in the process triggers safety risks", which is equivalent to the propagation path of internal elements of the HE.

Accordingly, we take the execution logic of experts in launching HAZOP as the ISK relationship in a novel way, i.e. $R_{ISK}(\zeta, \eta) = IC(\zeta, \eta) \to D(\zeta, \eta) \to ME(\zeta, \eta) \to C(\zeta, \eta) \to S(\zeta, \eta)$, which also enables ISK to circulate. For example, "液氨泄露(liquid ammonia leakage)" can be the IC of one HE or the C of another HE. See the left part of Fig. 2 and the brief presentation of Equ.8 based on Equ.7. It is worth noting that the $R_{ISK}(\zeta, \eta)$ is a strict standard during the implementation of HAZOP by experts, and also a certainty in the HAZOP report. Thus, $R_{ISK}(\zeta, \eta)$ can be applied to various processes in industry, even if their HAZOP descriptions are different. The ingenious design of $R_{ISK}(\zeta, \eta)$ greatly simplifies subsequent relationship extraction tasks, in the process of extracting entities from the HAZOP text, each HAZOP node description is taken as input, and only the elements in $R_{ISK}(\zeta, \eta)$ need to be filled in between ζ and η in turn. For example, see the right part of Fig. 3, where, HAZOP report 1 belongs to the aviation manufacturing process, for such a HAZOP node description, "员工误操作，氧含量过高，功率过低，发动机抖振，发动机损坏(staff mis-operation, too high oxygen content, too low power, engine chattering and engine damage)", its ISK system is constructed in this way: the corresponding relationship between 员工(staff, ζ) and 误操作(mis-operation, η) is IC, and the corresponding relationship between 氧含量(oxygen content, ζ) and 过高(too high, η) is D, and the rest are carried out in turn. Note that the extraction of entities such as 人员(staff) and 误操作(mis-operation) will be launched in the data layer (see Section 3.2). HAZOP report 2 belongs to the coal production process, for such a HAZOP node description, "管线存水，阀门开度过小，液位过低，压缩机喘振，压缩机损坏(pipeline water storage, too small valve opening, too low liquid level, compressor surge and compressor damage)", the construction of its ISK system is consistent with that described above.

$$E_{ss}(\zeta,\eta) = \{IC(\zeta,\eta) \to D(\zeta,\eta) \to \sum ME_i(\zeta,\eta) \to C(\zeta,\eta) \to S(\zeta,\eta)\} \tag{8}$$

**Table 1**
A brief representation in $E_{ss}(\zeta, \eta)$

| Element | Description | Representation |
|---|---|---|
| IC | 管线存水<br>(pipeline water storage) | ζ: 管线(pipeline)<br>η: 存水(water storage) |
| D | 压缩机冷后温度过高<br>(high temperature after compressor cooling) | ζ: 压缩机(compressor)<br>η: 冷后温度过高<br>(high temperature after cooling) |
| ME | 管线破裂<br>(pipeline rupture) | ζ: 管线(pipeline)<br>η: 破裂(rupture)<br>ζ: 压缩机(compressor)<br>η: 喘振(surge) |
| | 压缩机喘振<br>(compressor surge) | |
| C | 压缩机损坏<br>(compressor damage) | ζ: 压缩机(compressor)<br>η: 损坏(damage) |
| S | 温度控制器联锁<br>(temperature controller interlock) | ζ: 温度控制器<br>(temperature controller)<br>η: 联锁(interlock) |

In summary, we have formally completed the development of the ISKSF, which is the original of our research, including the construction of the $O_{ISK}$ and the design of the $R_{ISK}(\zeta, \eta)$. ISKSF is a general ISK system framework, which can standardize HAZOP reports in the same process or even in different processes, and uniformly represent diverse types of ISK. ISKSF is a practice of engineering design for industrial safety, it provides a practical scheme for the integration of ISK in the form of methodology the help of HAZOP, which widens the research of engineering design and extends the perspective for other researchers. We believe it can be used as a support for the self-improvement of engineering design.

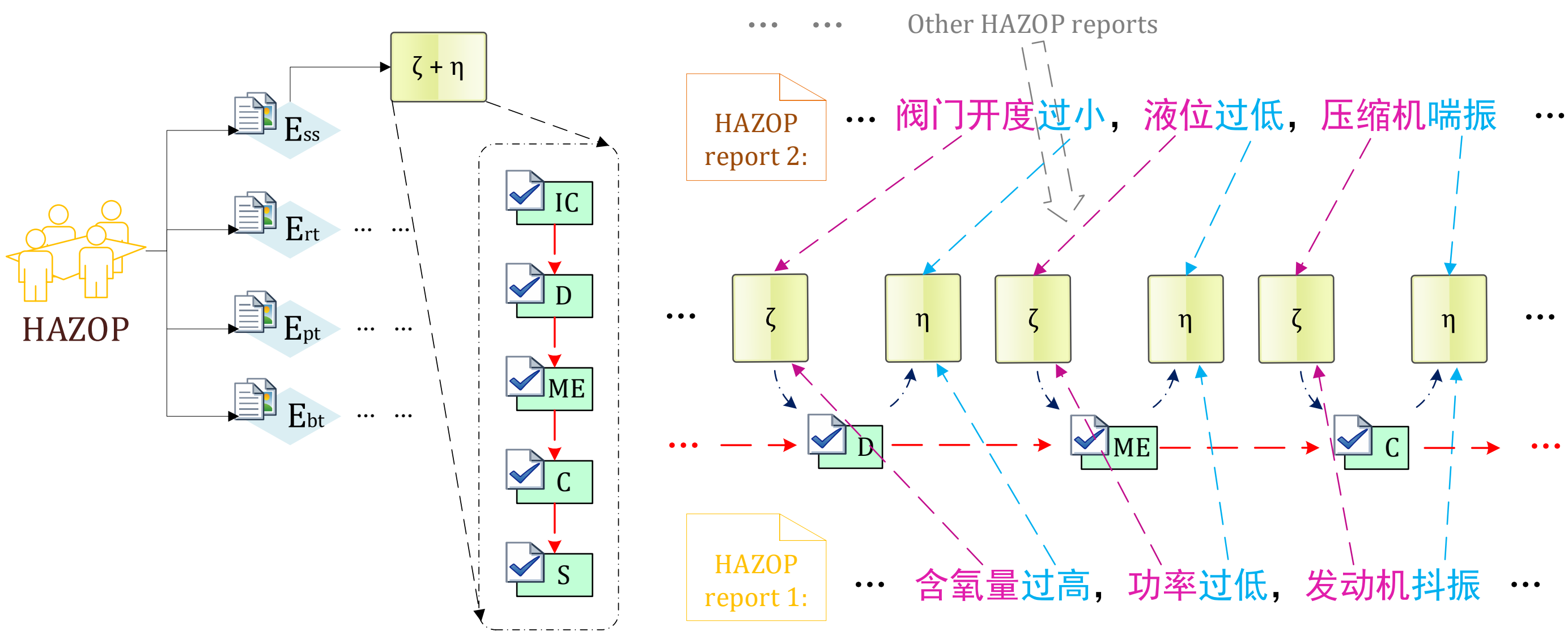

**Fig. 2.** The ISKSF (the left part) and its brief case (the right part)

### *3.2. Data layer*

The KG is integrated by triples which are in the form of < entity - relationship - entity >, thus, the data layer is composed of entity extraction and relationship extraction. In particular, the ISK relationship has been determined in the pattern layer, hence, this section is a detailed description of entity extraction. The HAZOP text mainly has the following particularities.

(1) There is a great deal of professional jargon, such as "富气 (rich gas)" and "反吹气 (blowback) ", etc.

(2) It is common that the same information discussed and recorded by different experts has different semantic expressions. For example, for guide words such as "高 (high)", "大 (large)" and "多 (many)", they have the same meaning frequently.

(3) The entity nesting is more diverse. For example, "乙二醇回收塔顶分液罐(glycol recovery tower overhead knockout drum)" is only a common equipment, which has four nested types: "乙二醇 (glycol)", "分液罐 (knockout drum)", "乙二醇回收塔(glycol recovery tower)" and "乙二醇回收塔顶分液罐(glycol recovery tower overhead knockout drum)".

More importantly, in terms of information structure, HAZOP reports conforming to the $R_{ISK}$(ζ, η) has strong logic. These particularities hinder the efficiency of information extraction. Therefore, it is a natural and critical thought to design an advanced information extraction model in industrial safety. We propose HAINEX with the help of text mining in data science, an information extraction model based on hybrid deep learning in the form of named entity recognition (NER), because NER in deep learning can effectively improve the performance of the model (Li et al., 2020). We will elaborate the data layer from three aspects: HAZOP dataset, HAINEX and experiments.

#### *3.2.1 HAZOP dataset*

Firstly, we obtained the HAZOP original text from the semi-structured and unstructured data such as professional websites, electronic reports and paper documents through preprocessing operations such as text conversion and text cleaning. Then, we collected 8655 HAZOP descriptions from the original text through the rule-based method. According to ζ and η in the pattern layer (Section 3.1), the material (such as "混醇(mixed alcohol)"), the equipment (such as "醇分离塔(alcohol separation tower)"), the state (such as "超压(overpressure)") and the consequence (such as "破裂(rupture)") in the HAZOP description were regarded as the ISK entity ($E_{ISK}$) to be extracted. In particular, some equipment is represented by a unique process label, for example, T-5753001 is the label of "醇分离塔(alcohol separation tower)", so the process label is also regarded as an $E_{ISK}$ to be extracted. Finally, we marked these entities in the format of "BIO" (Wang, Zhang and Gao; 2021), the concrete annotation strategy is shown in Table 2. The marked corpus with 86420 lines of data is the HAZOP dataset. which is divided into training set, validation set and test set in the ratio of 8:1:1.

**Table 2**
The concrete annotation strategy.

| Entity type | Initial | Subsequent |
|---|---|---|
| Equipment | B-EQU | I-EQU |
| Process label | B-PLA | I-PLA |
| Consequence | B-CON | I-CON |
| Material | B-MAT | I-MAT |
| State | B-STA | I-STA |
| Non-entity | O | O |

### 3.2.2 *HAINEX*

At present, the pre-training language model BERT (Devlin et al., 2018) has become the paradigm of NLP and shows excellent performance in information extraction tasks in various fields (Zhou et al., 2020; Nizzoli et al., 2020; Chen and Luo, 2019). BERT mainly achieves the purpose of extracting information semantic features from massive texts through two pre-training tasks: masked language model (MLM) and next sentence prediction (NSP), where, NSP is used to judge whether sentences are adjacent. However, the HAZOP information follows the $R_{\text{ISK}}(\zeta, \eta)$ (see the pattern layer), the fixed association between sentences has been satisfied, therefore, the benefit of NSP mechanism is not high, instead, it imposes a potential constraint that may treat HAZOP text more strictly, which leads to the omission of some semantic features by BERT, so that the recalled ISK is not comprehensive. In order to alleviate this problem, a natural idea is conceived in this research: we can regard the extracted semantic features as a candidate set ($C_{sf}$) and expand its cardinality to complete and enrich the features; Then, an auxiliary network further extracts features based on the $C_{sf}$. On the one hand, it is used to extract features belonging to other angles, on the other hand, it is similar to the operation of " screening " to reduce other unnecessary features that may be introduced due to the expansion of the $C_{sf}$; Finally, an appropriate loss function is used to enrich the recalled ISK as much as possible. Specifically, IBERT is proposed by optimizing BERT to expand and enrich semantic features; The auxiliary network used in this paper is BiLSTM, an encoder for obtaining the contextual information and long-distance dependence of information (Lindemann et al., 2021; Zhang, Yang and Zhou; 2021); CRF is used as a decoder to reduce label prediction errors and cooperate with the proposed loss function IL.

To sum up, HAINEX is composed of IBERT, BiLSTM and CRF&IL. Fig. 3 is the overall framework. The input of HAINEX is the HAZOP description, and the output is the label of the corresponding entity. Firstly, the joint embedding composed of the token embedding (a 768-dimensional random character embedding matrix in this paper), the segment embedding (a 20-dimensional embedding matrix mainly for NSP) and the position embedding (a 128-dimensional embedding matrix generated by random initialization to represent the position information of each character) generated by HAZOP descriptions is passed to IBERT; Secondly, IBERT transforms the joint embedding into a semantic vector containing semantic features and passes it to BiLSTM; Then, BiLSTM uses a pair of reverse LSTMs to encode the semantic vector to obtain the context vector with context features. Finally, CRF&IL decodes the context vector and outputs the tag sequence with the greatest probability. The following is a detailed description of each module.

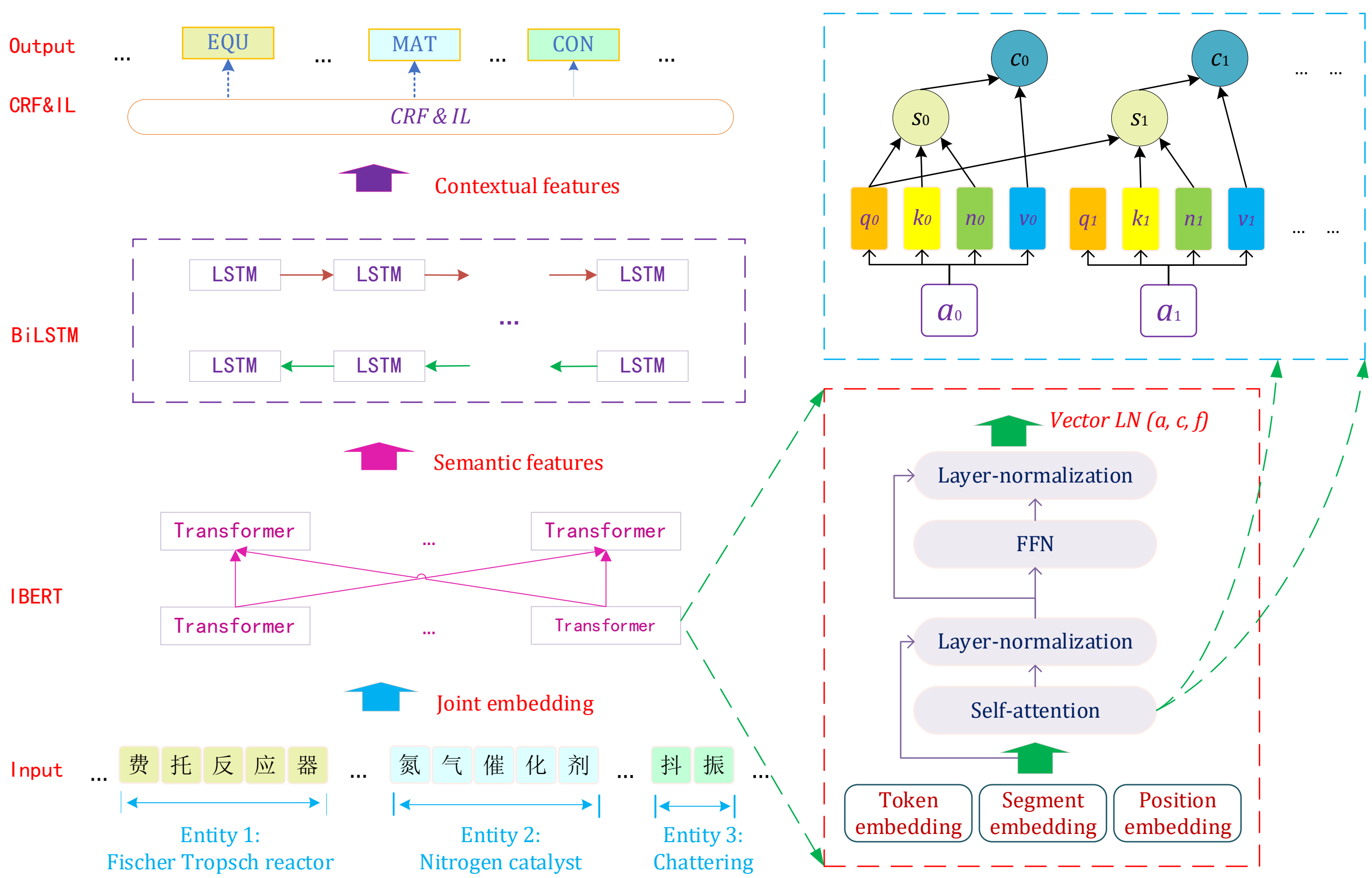

**Fig. 3.** The HAINEX model

*(1) IBERT module*

In the professional field, BERT can be fine-tuned to obtain the effect like transfer learning (Devlin et al., 2018). IBERT is an industrial safety-oriented variant of BERT, similar to BERT, IBERT is also superimposed by multiple encoders termed Transformer, the Transformer is shown in the red dotted box in Fig. 4. Firstly, the joint embedding $a = \{a_0, a_1, \ldots, a_n\}$ including the token embedding, the segment embedding and the position embedding is passed to the self-attention layer. Through the transformation capability of self-attention, the attention distribution $c = \{c_0, c_1, \ldots, c_n\}$ is obtained. Then, the vector $a$ and the vector $c$ are layer-normalized to obtain $LN(a, c)$, and the $LN(a, c)$ is transferred to the feedforward neural network (FFN) to obtain a deeper representation vector $f = \{f_0, f_1, \ldots, f_n\}$. Finally, the vector $f$ and the vector $LN(a, c)$ are layer-normalized to obtain the final output vector $LN(a, c, f)$. The self-attention is the core of Transformer, which eliminates the limitations of traditional RNN by weighted summation. The self-attention in BERT is shown in Equ.9.

$$c = v \cdot softmax(s(k,q)) \tag{9}$$

Where, $q = \{q_0, q_1, \ldots, q_n\}$ is the query vector, $k = \{k_0, k_1, \ldots, k_n\}$ is the key vector, $v = \{v_0, v_1, \ldots, v_n\}$ is the value vector. $q$, $k$ and $v$ are obtained respectively by different linear mappings of vector $a$, and their dimensions are the same. It is worth noting that $s(k, q)$ is the score function with scaling point product, as shown in Equ.10, where, $d_k$ is the dimension of $q$ or $k$.

$$s(k,q) = \frac{k^T q}{\sqrt{d_n}} \tag{10}$$

In order to enrich the $C_{sf}$ (see the beginning of 3.2.2), besides, compared with the general domain information, the HAZOP information is more complex and fuzzier (see the beginning of 3.2). Hence, in the IBERT, we weaken the strictness of extraction by giving self-attention some errors, which can help IBERT mine the semantic representation of HAZOP text more widely, make vector $c$ contain richer semantic features, and further alleviate the symptom of feature omission. As shown in the blue dotted box in Fig. 4, by adding a normal distribution bias to self-attention, we force IBERT to give some tolerance to HAZOP information, which can improve the generalization ability of IBERT, as shown in Equ.11 and Equ.12, where $n = \{n_0, n_1, \ldots, n_n\}$ is the bias vector. IBERT can alleviate the coupling of HAZOP information with its additional tolerance, and enrich the extracted semantic features.

$$c_{HAINEX} = v \cdot softmax(s(k,q,n)) \tag{11}$$

$$s_{HAINEX}(k,q,n) = s(k,q) + \frac{n}{\sqrt{d_n}} = \frac{k^T q + n}{\sqrt{d_n}} \tag{12}$$

*(2) BiLSTM module*

LSTM is a special RNN for processing long time series data, which is mainly composed of three parts described below. Note that for the subscript of the variable in this section, "$t$" represents the current time step, and "$t$-1" represents the previous time step. The first part is information screening, the input value $x_t$ is accompanied by a coefficient $f_t$ with an interval of [0,1], the $f_t$ is the mapping of *sigmoid* function based on the $x_t$ the output value $h_{t-1}$. If the $f_t$ is 1, the whole $x_t$ will be retained. If the $f_t$ is 0, the whole $x_t$ will be discarded, as shown in Equ.13. The second part is information addition. Firstly, the $h_{t-1}$ and $x_t$ are simultaneously *tanh* functionalized to generate the candidate vector $C_t$*. Then, the $h_{t-1}$ and the $x_t$ are simultaneously *sigmoid* functionalized to generate the weight $i_t$, which can regulate the update of $C_t$*. Finally, the state vector $C_{t-1}$ and the $C_t$* are superimposed to update the $C_t$, as shown in Equ.14. In the third part, the $h_{t-1}$ and $x_t$ are simultaneously *sigmoid* functionalized to generate the weight $k_t$, the $k_t$ interacts with the functionalized $C_t$ to obtain $h_t$, as shown in Equ.15. These operations enable LSTM to have memory ability and retain only important features, which can alleviate the problems of the poor long-term dependence, the gradient vanishing and the gradient explosion. BiLSTM is composed of the forward LSTM and the backward LSTM, it can encode the head and tail of the text sequence respectively to capture the context features with the long-distance dependencies.

Therefore, as the auxiliary network of HAINEX, BiLSTM can further extract contextual features from the semantic vector output by IBERT. Additionally, since features do not exist in isolation, BiLSTM can also play a similar role of " screening " while mining context features to reduce unnecessary features that may be introduced by IBERT while expanding the cardinality of $C_{sf}$.

$$f_t = sigmoid(\omega_f \cdot [h_{t-1}, x_t] + b_f) \tag{13}$$

$$\begin{cases} i_t = sigmoid(\omega_i \cdot [h_{t-1}, x_t] + b_i) \\ C_t* = tanh(\omega_C \cdot [h_{t-1}, x_t] + b_C) \\ C_t = f_t \cdot C_{t-1} + i_t \cdot C_t * \end{cases} \tag{14}$$

$$\begin{cases} k_t = sigmoid(\omega_k \cdot [h_{t-1}, x_t] + b_k) \\ h_t = k_t \cdot tanh(C_t) \end{cases} \tag{15}$$

*(3) CRF&IL module*

For BiLSTM, each label is regarded as a separate existence, the maximum probability of labels is paid more attention, and the dependency between labels is not considered, which causes the predicted information to be chaotic. To avoid this limitation, CRF is introduced as the decoder in HAINEX. The relationship between adjacent labels can be considered by CRF to obtain the global optimal prediction sequence. The maximum likelihood estimation loss function of CRF is shown in Equ.16.

$$Loss(x, y', y) = log \sum_{y'} e^{s(x,y')} - s(x,y) \tag{16}$$

Where $x = \{x_0, x_1, ..., x_n\}$ is the input sequence, $y' = \{y_0', y_1', ..., y_n'\}$ is the prediction sequence, $y = \{y_0, y_1, ..., y_n\}$ is the real sequence, and $s(x, y)$ is the sum of the emission score and the transition score (Lample et al., 2016). However, NSP mechanism limited by HAZOP text properties may impede the process of HAZOP information extraction, which makes the recalled ISK incomplete (see the beginning of 3.2.2). Therefore, in order to cooperate with the above two modules and further enrich the recalled ISK as much as possible, an intuitive idea is that we appropriately weaken the strictness of CRF loss function, and make the prediction ability of CRF fluctuate by adjusting $y'$. This improved loss function for industrial safety is termed IL. We support and encourage this fault tolerance of IL, which makes the model fit in a more divergent direction and recall more ISK, as shown in Equ.17 - 19.

$$IL = Loss_{HAINEX}(x, f(y), y) = log\sum_{f(y)} e^{s(x, f(y))} - s(x, y) \tag{17}$$

$$f(y) = \begin{cases} \alpha y' + |N_\alpha|, \tau > s \\ \beta y' - |N_\beta|, \tau \le s \end{cases} \tag{18}$$

$$\tau = f_\tau(y') \tag{19}$$

Where, $s$ is the set threshold, which is a trainable value in the interval [0,1], $\tau$ is the mapping value of the original prediction value, the prediction sequence is adjusted according to the value of $\tau$. If $\tau$ greater than $s$, a coefficient $\alpha$ is given to the original sequence, and a disturbance sequence $N_\alpha$ ($N_\alpha > 0$) is appended, otherwise, a coefficient $\beta$ is given to the original sequence, and a disturbance sequence $N_\beta$ ($N_\beta > 0$) is removed, in this research, the *sigmoid* function is used to map $y'$ to $\tau$. The hyperplane of HAZOP label classification can be appropriately optimized by operations above, which makes the HAINEX more generalized to industrial safety and expands the diversity of extracted ISK.

### 3.2.3. ***Experiments***

Large quantities of test experiments show that when $\alpha = 1.15$, $\beta = 1$, the performance of the HAINEX is the optimal. Therefore, in this research, α and β are set to 1.15 and 1, respectively. The indicators (Li et al., 2021) of the precision (P), the recall (R) and the F1-score (F1) are considered to evaluate the extraction performance of the HAINEX. More importantly, the common hyperparameters in all neural networks are the same for both evaluation experiments and comparison experiments followed. For example, we apply the Adam optimizer with the learning rate 1e-3, the ReLU activation function, the batch size with the value 64 and the BERT-base (for networks with BERT) with 12 hidden layers and 768 hidden size. Besides, we train 50 epochs for all models in both the validation set (Val) and the test set (Test). The results of the evaluation experiment are shown in Table 3. It can be shown that the HAINEX has strong extraction performance. Where, the F1-score of the total entity on the test set and validation set reached 88.37% and 88.55% respectively. The precision, the recall and the F1-score of "Process label" and "Consequence" exceeded 94% on the two set, and the recall of "Process label" even reached 100% on the test set. The results show the feasibility of HAINEX in the ISK extraction task.

**Table 3**
The results of evaluation experiments

| Model | Entity type | P (%) | | R (%) | | F1(%) | |
|---|---|---|---|---|---|---|---|
| | | Test | Val | Test | Val | Test | Val |
| HAINEX | Total | 88.42 | 88.59 | 88.32 | 88.52 | 88.37 | 88.55 |
| | Equipment | 84.23 | 83.99 | 84.48 | 84.36 | 84.35 | 84.17 |
| | Process label | 97.32 | 98.59 | 100.00 | 99.82 | 98.24 | 99.20 |
| | Material | 80.00 | 78.65 | 74.58 | 73.22 | 77.19 | 75.84 |
| | State | 86.59 | 86.27 | 84.86 | 84.33 | 85.71 | 85.29 |
| | Consequence | 94.44 | 94.84 | 96.23 | 95.97 | 96.63 | 95.73 |

Table 4 is the results of comparison experiments, including common NER models, such as CRF (Peng and Andrew, 2006), IDCNN-CRF (Strubell et al., 2017), BiLSTM-CRF (Huang, Xu and Yu, 2015), SelfAtt-IDCNN-CRF (Huang et al., 2020), BiLSTM-IDCNN-CRF (Fang et al., 2020), Att-BiLSTM-CRF (Luo et al., 2018), BERT-CRF (Souza, Nogueira and Lotufo, 2020), BERT-BiLSTM-CRF (Li, Zhang and Zhou, 2020). In addition, models (number 7-12) are experiments to further verify the effectiveness of the proposed optimization idea (alleviating the potential limitations of NSP). Where, the "IDCNN" means the dilated CNN (Yu and Koltun, 2016), a common convolutional neural network structure for NER that can capture edge features; the "SelfAtt" is short for the self-attention mechanism, and the "Att" is short for the attention mechanism (Vaswani et al., 2017).

The results proved that the HAINEX achieves the optimum information extraction performance. No matter in the test set or the validation set, the precision of the existing models (number 1-8) has not reached the level of 86%, and the F1-score is also far behind HAINEX. For F1-score on the test set and the validation set, HAINEX is 1.70 and 1.74 percentage points higher than the original BERT-BiLSTM-CRF (model 8), the model 11 is 1.24 and 0.75 percentage points higher than the model 9 (the proposed method, IBERT-CRF&IL, is applied to the auxiliary network IDCNN), and model 10 is 0.70 and 0.64 percentage points higher than model 7 (without the auxiliary network). The above improvements show the effectiveness of the proposed method, which can alleviate the constraints of NSP and improve the performance of the model. It can be further found that for the proposed method itself, the improvement without auxiliary network is weaker than that with auxiliary network. Besides, for the auxiliary network (model 11-12), the improvement effect of BiLSTM is better than that of IDCNN, which indicates that the selection of auxiliary network is worth considering, compared with the edge feature extraction of IDCNN, BiLSTM is more suitable for screening the semantic vector output by IBERT in the form of capturing contextual features, we speculate that the semantic vector may contain more edge features.

**Table 4**
The results of comparison experiments (for the total entity)

| Number | Model | P (%) | | R (%) | | F1(%) | |
|---|---|---|---|---|---|---|---|
| | | Test | Val | Test | Val | Test | Val |
| 1 | CRF | 75.45 | 74.33 | 76.60 | 74.99 | 76.02 | 74.66 |
| 2 | IDCNN-CRF | 83.26 | 83.94 | 83.65 | 84.02 | 83.45 | 83.98 |
| 3 | BiLSTM-CRF | 82.20 | 82.17 | 84.24 | 84.88 | 83.21 | 83.50 |
| 4 | Att-BiLSTM-CRF | 83.85 | 84.66 | 84.81 | 84.23 | 84.33 | 84.44 |
| 5 | SelfAtt-IDCNN-CRF | 84.59 | 85.63 | 83.88 | 84.35 | 84.23 | 84.99 |
| 6 | BiLSTM-IDCNN-CRF | 84.01 | 83.65 | 85.22 | 85.03 | 84.61 | 84.33 |
| 7 | BERT-CRF | 85.49 | 84.83 | 86.55 | 85.94 | 85.75 | 85.68 |
| 8 | BERT-BiLSTM-CRF | 85.38 | 85.95 | 87.99 | 87.68 | 86.67 | 86.81 |
| 9 | BERT-IDCNN-CRF | 85.60 | 85.92 | 86.83 | 86.43 | 86.21 | 86.17 |
| 10 | IBERT-CRF&IL | 85.88 | 85.26 | 87.02 | 86.35 | 86.45 | 85.80 |
| 11 | IBERT-IDCNN-CRF&IL | 86.94 | 86.77 | 87.96 | 87.08 | 87.45 | 86.92 |
| 12 | HAINEX | 88.42 | 88.59 | 88.32 | 88.52 | 88.37 | 88.55 |

All in all, HAINEX is advanced and reliable in the task of extracting ISK. The proposed method can alleviate the constraint of NSP on the HAZOP report with strong structure and improve the adaptability of BERT in industrial safety engineering. Except for the HAZOP report recording massive ISK, other highly structured and logical industrial information containing domain knowledge is also common, HAINEX can provide optimization ideas and feasible schemes for the extraction of these knowledge, and sets an example for the practical application (Chiarello et al., 2019) of data science (such as text mining and information extraction) for engineering design. We hope it can inspire other researchers who are committed to data science for engineering design and give full play to the expected value.

### *3.3. Storage layer*

We construct ISK triples in the form of Equ.20 between the extracted $E_{ISK}$ (see Section 3.2) and the set relationship $R_{ISK}(\zeta, \eta)$ (see Section 3.1). Where the set of the ISK triples $<n_h, e, n_t>$ are composed of a node head $n_h$ and a node tail $n_t$ in which both belong to the entity $E_{ISK}$, both nodes are linked together using an edge $e$ that represents the $R_{ISK}(\zeta, \eta)$ that lies between the two entities. Specifically, with reference to Fig. 2, each HAZOP node description is taken as the input, the entities (ζ and η) in the input are extracted by HAINEX, and the elements in $R_{ISK}(\zeta, \eta)$ are successively embedded between ζ and η. Considering that there may be more than one ME, we firstly construct triples about IC, D and S from both ends of the input, and then construct triples about ME.

$$ISKG = \{< n_h, e, n_t > | n_h, n_t \in E_{ISK}; e \in R_{ISK}\} \tag{20}$$

Fig. 4 is an illustration of the formed ISK triples about the description of "PICS0501 fault, the pressure of T-5642103 is too low, the rich liquid flows in and the fuel gas pipe network is damaged, it is recommended to add PV0501 interlocking". Where the PICS0501 is a kind of pressure controller, the T-5642103 is a kind of rich liquid flash tank and the PV0501 is a kind of pressure regulating valve.

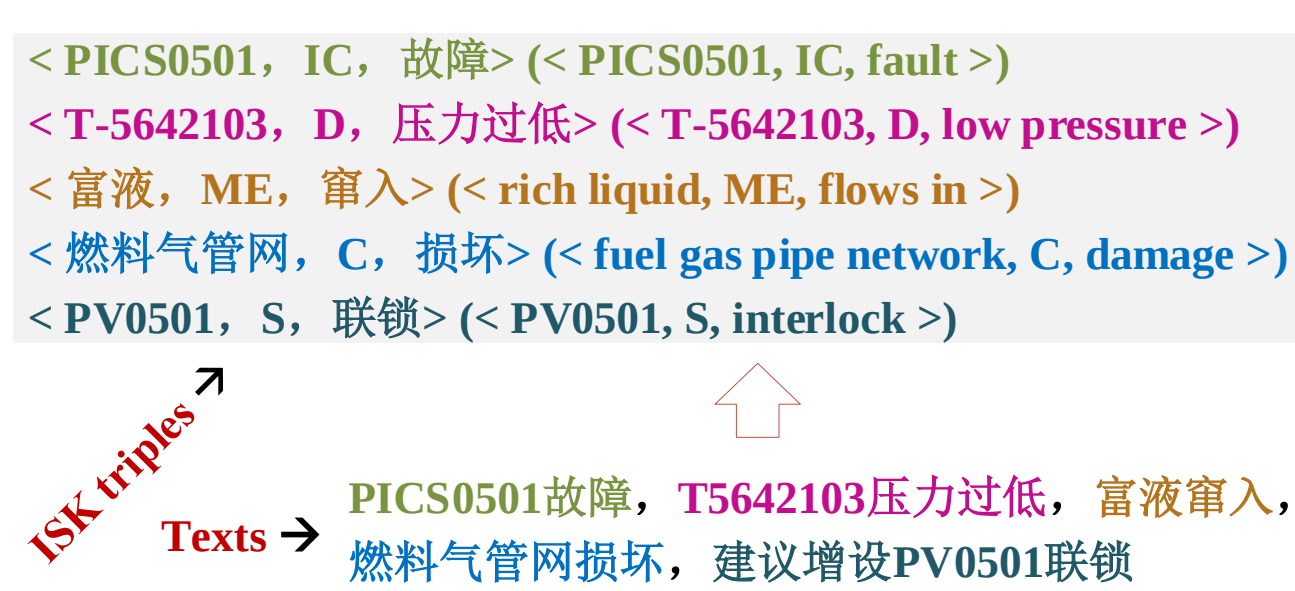

**Fig. 4.** An illustration of the formed ISK triples

We import ISK triples into the Neo4j graph database (Miller, 2013; Huang et al. 2020; Kim et al., 2021) to accomplish the uncanonicalized ISKG. Due to the complexity of the process, some entities and relationships are redundant, we leverage the rule matching to eliminate them, such as merging the same elements. It is worth noting that HAZOP reports record the results with standardization and preciseness of multiple analysis by the expert group, even if the literal meaning of words is very close, each word still has a unique meaning, such as "工艺管线 (process pipeline)" and "管线(pipeline)". Therefore, it is not necessary to solve the problem of ambiguity at present, and entity disambiguation in industrial safety engineering is considered a research problem on its own that is out of the scope of this paper. In future work, we will incorporate other various industrial safety reports such as failure mode and effects analysis reports (Kiran, 2017) and fault tree analysis reports (Swapan, 2017) into ISKG, at that time, the operations, such as entity fusion and entity disambiguation, will be launched.

## 4. CASE STUDY

In this study, we cooperated with Shenhua Group, which ranked 165th among the world's top 500 enterprises in 2014, to study ISKG for 4 million T / a coal indirect liquefaction process. This process is one of the largest coal energy projects in the world, which is of great significance in coal resource utilization, sustainable energy development and oil resource substitution. Two procedures are mainly included in the process, one is the gasification, under the high temperature condition, the syngas (the main component is a mixture of carbon monoxide and hydrogen) is generated by the reaction of the coal, the oxygen and the steam, the other is the liquefaction. The syngas is combined into the liquid fuel under the action of the catalyst. This process consists of nine units, the brief introduction of each unit is shown in Table 5. We collected 834 analysis records from 102 nodes to develop ISKG. Based on AE-5611101 (a kind of oil gas air cooler), Fig. 5 shows a uncanonicalized ISKG fragment. Fig. 6 illustrates the canonicalized version of Fig. 5, where some nodes and edges are merged after the rule matching process (see Section 3.3).

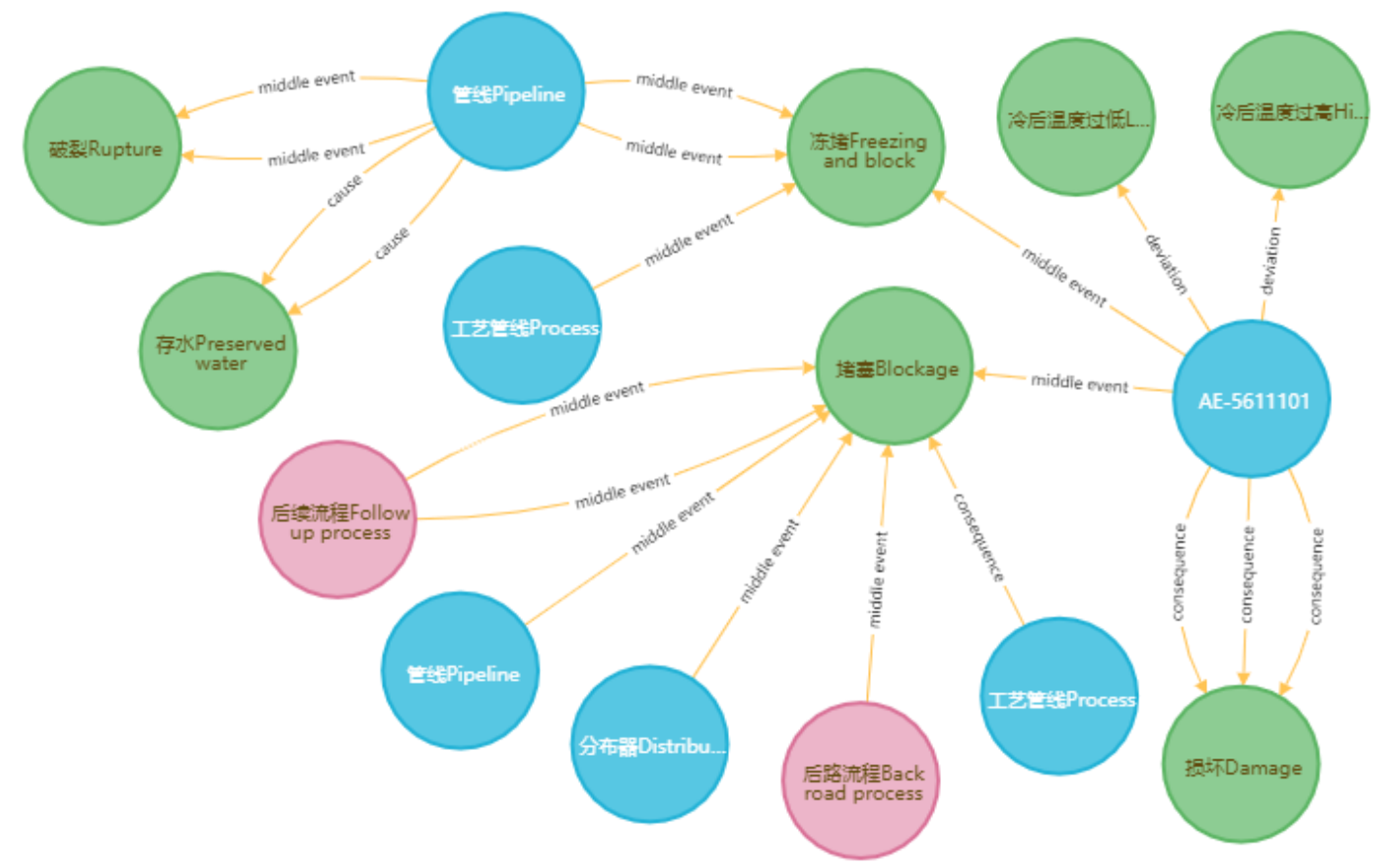

**Fig. 5.** The uncanonicalized ISKG fragment

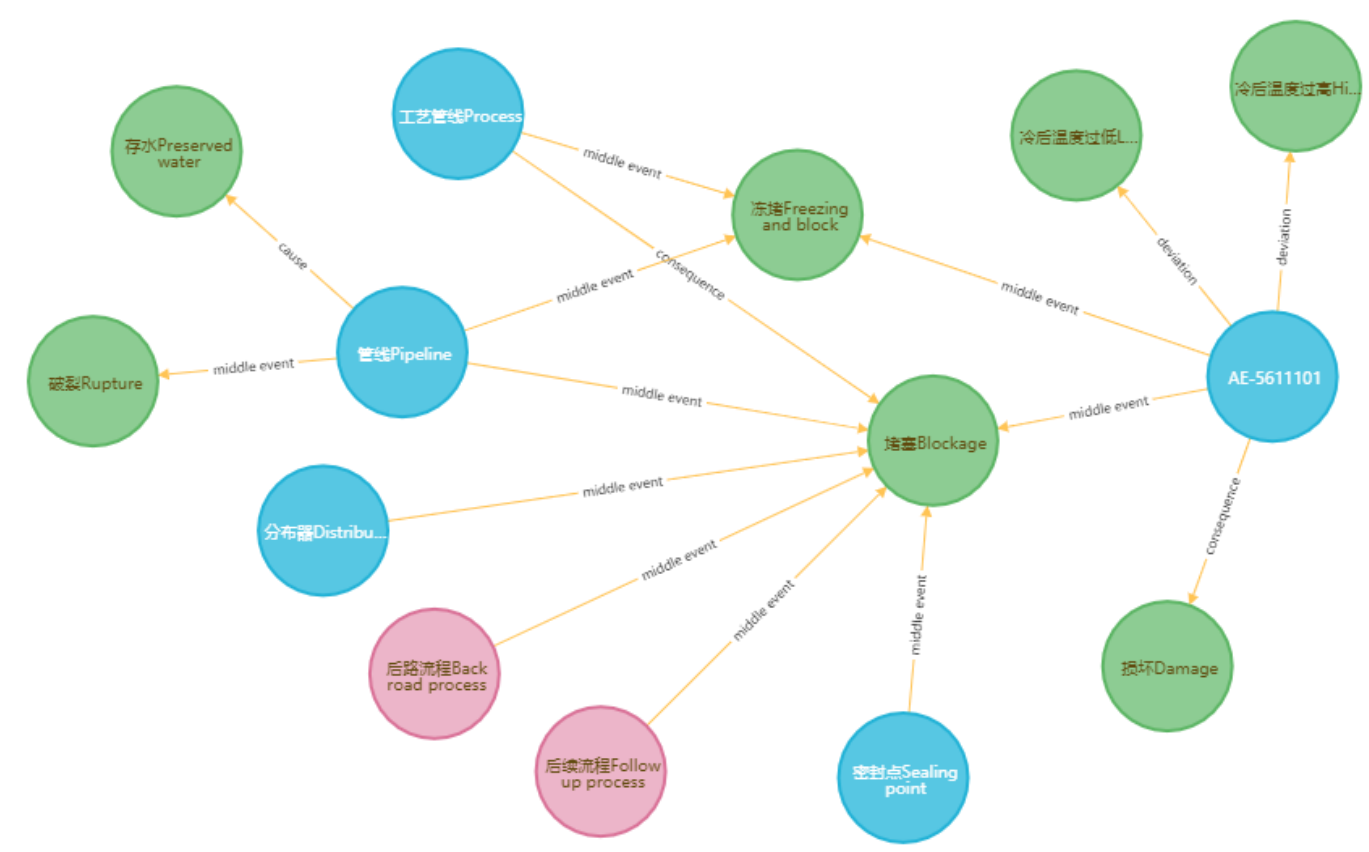

**Fig. 6.** The canonicalized ISKG fragment

**Table 5**
The brief introduction of indirect coal liquefaction process

| Unit | Brief description | Main device |
|---|---|---|
| Fischer Tropsch synthesis | The catalyst slurry is obtained by reduction reaction of the heavy diesel, the catalyst and the circulating gas | Reduction reactor |
| Catalyst reduction | The discharge and feed of the stripper exchange heat, and the removed heat source enters the stripper | Heavy oil stabilized wax heat exchanger |
| Wax filtration | The stabilized wax from the Fischer Tropsch synthesis unit is received and sent to the subsequent unit after boosting | Stabilizing wax tank |
| Tail gas decarbonization | The decarburized gas enters the compressor after being separated | Decarbonization gas knockout drum |
| Fine desulfurization | The sulfur content of syngas is reduced, and the purified syngas is sent to the subsequent unit | Fine desulfurization heater |
| Synthetic water treatment | The cooling water, the domestic water, the generated water and the demineralized water are provided | Pipeline |
| Liquid intermediate raw material tank farm | The storage of the intermediate raw materials | Oil wash naphtha pump |
| Low temperature oil washing | The cooled decarburized purified gas is received after water washing and purification | Purified gas separator |
| Deaerated water and Condensate refining station | The residual oxygen in demineralized water is removed to provide qualified reactants for subsequent processes | Built-in deaerator |

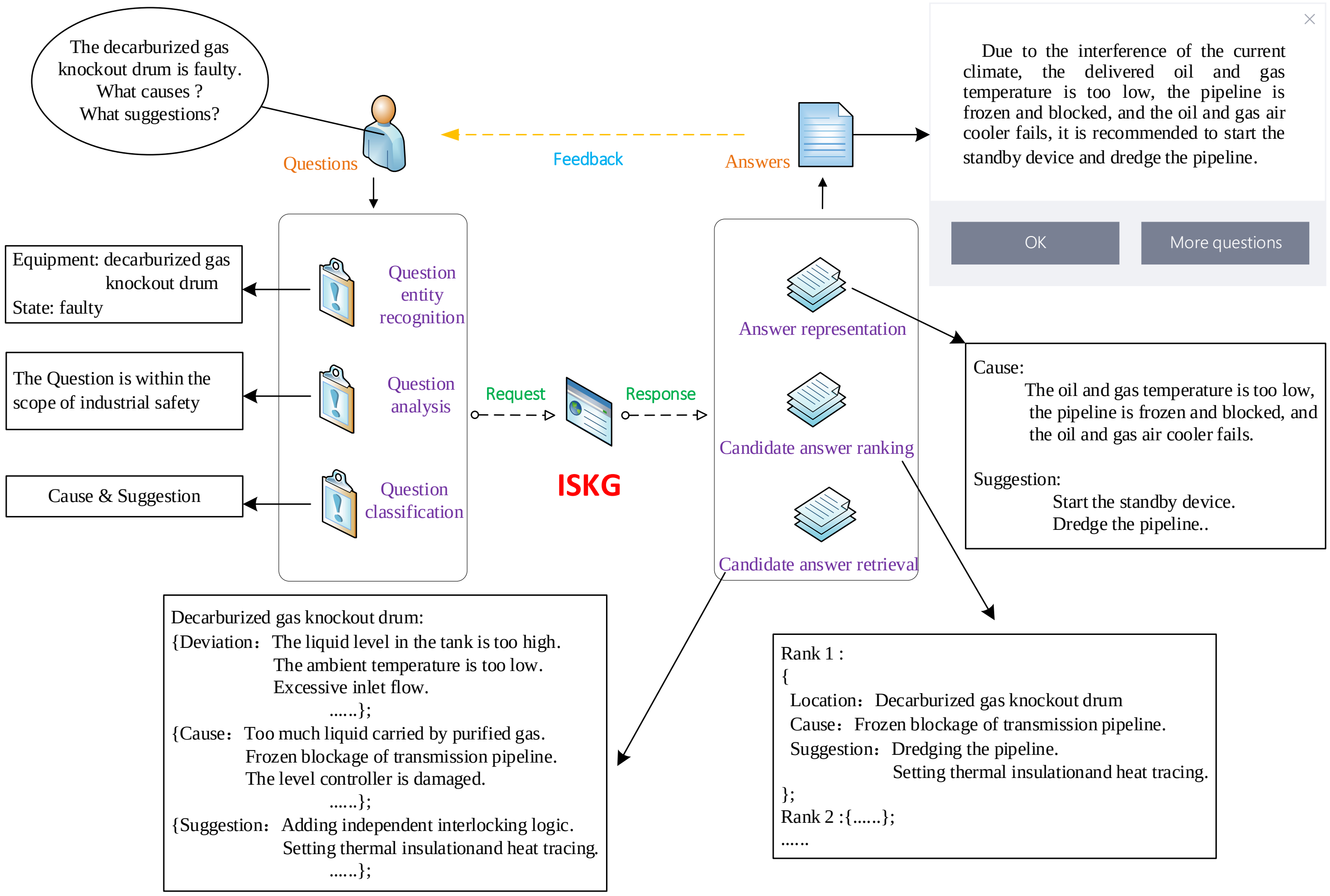

**Fig. 7.** The QAS, where, the left part is the request module and the right part is the response module

ISKG oriented applications, such as ISK visualization, ISK retrieval, auxiliary HAZOP and hazard propagation reasoning, can reuse, share and stimulate the value of ISK, and further improve HAZOP execution efficiency, which is of great significance to improve the security of the system. They can be used as the support for the integration of engineering design and data science into industrial safety. The following is a brief introduction.

(1) ISK visualization. We select the node AE-5611101 in Fig.5 as an example, in the tail gas decarbonization module, the upstream light oil and gas are separated into the light oil and the water under the cooling of AE-5611101 to provide satisfactory circulating gas for the subsequent compressor. ISKG takes the "冷却后温度(temperature after cooling)" as the starting point of hazard analysis, and clearly presents the deviation relationship between "AE-5611101" and "冷却后温度过高(too high temperature after cooling)" and "冷却后温度过低(too low temperature after cooling)". The cause for the freezing and blocking of pipeline is the water storage, which is also intuitively emerged. Besides, the middle event can also be described methodically, "管线冻堵(pipeline freezing and blocking)" or "管线堵塞(pipeline blockage)" is accompanied by "管线破裂(pipeline rupture)". Moreover, ISKG intuitively indicates that the consequence of AE-5611101 caused by these risks is "损坏(damage)". The ISK visualization reduces the complexity of hazard analysis, assists staff to master safety knowledge more conveniently, and improves the accuracy of hazard investigation, the completeness of analysis results and the safety of process.

(2) ISK retrieval. The ISK we need can be effectively located to present relevant information for us. For example, when we retrieve the C-5611101 (a circulating gas compressor), its potential hazards (such as overpressure and surge), cooperative equipment (such as Fischer Tropsch synthesis reactor and Fine desulfurization heater), related materials (such as blowback gas and saturated steam) and suggestions (such as interlock and compressor shutdown) can be provided by ISKG, which is convenient for the staff to comprehensively master ISK and be familiar with process flow and related operation.

(3) Auxiliary HAZOP. Hazard analysis can be executed based on fault location. When accidents occur, it is troublesome for operators to obtain reasonable solutions without fully understanding the process procedures because the process is too complex. In view of this problem, according to the abnormal phenomena, operators can locate node deviations in hazard events through ISKG and trace back fault causes, so that suggestions can be put forward quickly and hidden dangers can be eliminated effectively, which is of great significance to further improve the safety of the system.

(4) Hazard propagation reasoning. Omissions in HAZOP are inevitable. To further improve HAZOP, we can realize risk reasoning through ISKG to revise and make up for HAZOP. ISKG can infer other possible propagation paths in hazard events according to paths between existing entities, which makes the whole risk analysis of the process more perfect. For example, blockage may be a middle event of one entity or a consequence of another entity. Ammonia leakage may be a consequence in one hazard or a cause in another hazard. It can assist experts in brainstorming to optimize HAZOP.

More importantly, the question-answering system (QAS) based on ISKG is shown in Fig. 7. It is worth noting that the QAS is currently primary (based on a series of rule templates), which is more for the convenience of popularizing safety knowledge and enhancing prevention awareness for non-professionals, it will be optimized to be applied to the actual industry as a product in the future. Like the conversational system studied by Coli et al. (2020), the QAS can also be supported for industrial maintenance.

The QAS is mainly divided into the request module and the response module. In the request module, when a question is raised by users. Firstly, QAS will extract the entities in the question, and then analyze the entities to judge the domain scope of the question. If it is within the scope of industrial safety, QAS will determine the type of answer in combination with these entities and other keywords in the question, and request ISKG based on this information; Otherwise, QAS directly provides users with feedback that the scope of the question is not in line with industrial safety. In the response module, the candidate answers matched by ISKG are retrieved, sorted and represented by QAS (Soares and Parreiras, 2020), and Top-k appropriate answers are fed back to the user. For example, for the question description as " 油气空冷器发生故障，故障原因是什么？如何处理？ (The decarburized gas knockout drum is faulty. What causes? What suggestions?) ". Firstly, HAINEX extracts the relevant entities in the question, i.e. "Equipment: decarbonization gas knockout drum; State: faulty"; QAS judges that the question conforms to the field of industrial safety, and combines these entities and other keywords in the question to correspond the answer to the "cause" and the "suggestion". Then, QAS will retrieve a series of candidate answers composed of elements such as specific deviation, possible causes and feasible suggestions from ISKG, and sorts them. Finally, Top-k appropriate answers will be encapsulated and fed back to the user, for example, the Top-1 answer can be expressed as: "受目前的气候干扰，输送来的油气温度过低，管道冻堵，油气空冷器发生故障。建议启用备用装置，并疏通管道 (Due to the interference of the current climate, the delivered oil and gas temperature is too low, the pipeline is frozen and blocked, and the oil and gas air cooler fails, it is recommended to start the standby device and dredge the pipeline)". Compared to the search engine, the QAS of this study is professional, the user's requirements can be understood, it provides more straightforward answers and more expertise rather than displaying web links and information sequences. QAS can not only assist professionals in improving work efficiency, but also conduct safety teaching demonstration in the classroom to popularize the safety knowledge and enhance prevention awareness for non-professionals.

All in all, with the help of data science, ISKG and its application liberate the HAZOP report, realize the reuse and excitation of ISK, and optimize HAZOP, they can be used as the support for the integration of engineering design and data science in industrial safety. We hope they can inspire other researchers who are committed to data science for engineering design and give full play to the great value.

## *5. CONCLUSION*

HAZOP is a remarkable representative in industrial safety engineering, which can provide safety analysis decision-making for any process in the industry. However, the rich industrial safety knowledge (ISK) contained in HAZOP reports has not been thoroughly exploited. In order to reuse and unlock the value of ISK and optimize HAZOP, we have developed a novel knowledge graph with HAZOP as the carrier (ISKG) through bridging data science and engineering design.

ISKG follows the knowledge graph based approach. Specifically, firstly, in view of the dilemma of different expression of ISK, we have creatively developed a knowledge standardization framework (ISKSF) by deconstructing and generalizing HAZOP in a top-down manner. ISKSF contains the ontology and relationship of ISK, which can standardize HAZOP reports in various processes, uniformly represent and integrate ISK. Moreover, ISKSF also greatly simplifies relationship extraction tasks. It is a practice of engineering design for industrial safety. Secondly, considering the particularity of HAZOP text, we have skillfully conceived an information extraction model termed HAINEX based on deep learning in combination with data science, it can extract the ISK in HAZOP reports based on the ISKSF, which is a realization of data science for engineering design. Briefly, HAINEX consists of a variant of a pre-training language module, the bidirectional long short-term memory network and a conditional random field with an improved loss function. Finally, we build ISK triples based on ISKSF and HAINEX and store them in the Neo4j graph database.

We have launched ISKG for the indirect coal liquefaction process. ISKG is an outstanding example of the interaction between data science and engineering design, which can integrate and unlock the value of the HAZOP report in a feasible way, and its oriented applications, such as ISK visualization, ISK retrieval, auxiliary HAZOP and hazard propagation reasoning, can reuse, share and stimulate the value of ISK, and further improve HAZOP execution efficiency, which is of great significance to improve the security of the system. More importantly, the question-answering system can popularize safety knowledge and enhance prevention awareness for non-professionals. These applications that inject data science into engineering design can be used as the support for the industrial development.

In future work, we will incorporate other various industrial safety reports such as failure mode and effects analysis reports and fault tree analysis reports, and supplement additional tasks such as entity disambiguation and entity fusion. All in all, we foresee that in the near future ISKG will become sufficiently mature to provide added value to daily practices in industrial safety and enlighten other researchers committed to engineering design for data science.